\documentclass{article}

\usepackage{arxiv}
\usepackage[utf8]{inputenc} 
\usepackage[T1]{fontenc}    
\usepackage{float}
\usepackage{hyperref}       
\usepackage{url}            
\usepackage{booktabs}       
\usepackage{amsfonts}       
\usepackage{nicefrac}       
\usepackage{microtype}      
\usepackage{lipsum}
\usepackage{bbm}
\usepackage{graphicx, caption, subcaption, hyperref}
\usepackage{amsmath}
\usepackage{longtable}
\usepackage[utf8]{inputenc}
\graphicspath{ {./images/} }

\title{GIMP-ML : Python Plugins for using Computer Vision Models in GIMP}

\author{Kritik Soman\\
	Department of Electrical Engineering, \\
	Indian Institute of Technology Kanpur, \\
	India\\
	\texttt{kritiksoman@ieee.org}
}

\begin{document}
	\maketitle
	\begin{abstract}
		This paper introduces GIMP-ML v1.1, a set of Python plugins for the widely popular GNU Image Manipulation Program (GIMP). It enables the use of recent advances in  computer vision to the conventional image editing pipeline. Applications from deep learning such as monocular depth estimation, semantic segmentation, mask generative adversarial networks, image super-resolution, de-noising, de-hazing, matting, enlightening and coloring have been incorporated with GIMP through Python-based plugins. Additionally, operations on images such as k-means based color clustering have also been added. GIMP-ML relies on standard Python packages such as \texttt{numpy, pytorch, open-cv, scipy.} Apart from these, several image manipulation techniques using these plugins have been compiled and demonstrated in the YouTube channel (\url{https://youtube.com/user/kritiksoman}) with the objective of demonstrating the use-cases for machine learning based image modification. In addition, GIMP-ML also aims to bring the benefits of using deep learning networks used for computer vision tasks to routine image processing workflows. The code and installation procedure for configuring these plugins is available at \url{https://github.com/kritiksoman/GIMP-ML}.
	\end{abstract}


	\section{Introduction}
	
	Image editing has conventionally been performed manually by users or graphics designers using various image processing tools or software. A plethora of image editing and transformation functions are provided in such tools, which are available in open-source, commercial or proprietary license-based modes. Image processing workflows have varying levels of complexity and sometimes even require significant effort from the user even for simple modifications to images.

	GNU Image Manipulation Program (GIMP) is a popular free and open source image editing software that has been widely used on Linux-based platforms, as well as on other operating systems. It provides several features for image editing and manipulation and has a simple user interface to work with. It also supports the development of plugins which can be developed independently and integrated with the local GIMP installation on a computer. Using plugins, one can realize custom workflows or set of operations that can be applied to an image.
	
	Recently, machine learning techniques have completely changed the landscape of image understanding and many applications which were previously not possible have now become the new baseline. This has significantly been facilitated by recent advances in deep learning and the applications of resultant models to tasks in the computer vision domain. However, these deep learning models have been made available to users using independent deep learning frameworks such as Keras, TensorFlow, PyTorch, among others. It may also be noted here that since these networks have a \textit{``large''} architecture, their training is done on compute-intensive platforms (using GPUs) and the resultant models have a high memory footprint. Since the use of these models requires the user to code, graphics designers and users involved in conventional image editing workflows using image processing tools have not often been able to directly leverage the benefits from the deep learning models. As such, developing a framework that would enable the use of deep learning models in image editing tasks through commonly available image processing tools would potentially benefit both the deep learning / computer vision community as well as graphics designers and common users of such software.
	
	The motivation for this paper is to bridge the gap between cutting edge research in deep learning (computer vision) and manual image editing, specifically for the case of GIMP. A pilot implementation of plugins for GIMP, collectively termed as ``GIMP-ML'' (GIMP - Machine Learning), have been presented for various tasks such as background blurring, image coloring, face parsing, generative portrait modification, monocular depth based relighting, motion blurring, image de-noising, de-hazing, matting, enlightening and generating super-resolution images. It is expected that the image editing process would become highly automated in the upcoming future as the semantic understanding of images improves, which would be facilitated by advances in artificial intelligence.
	
	\begin{figure}[H]
    \centering
	\includegraphics[width=1\textwidth]{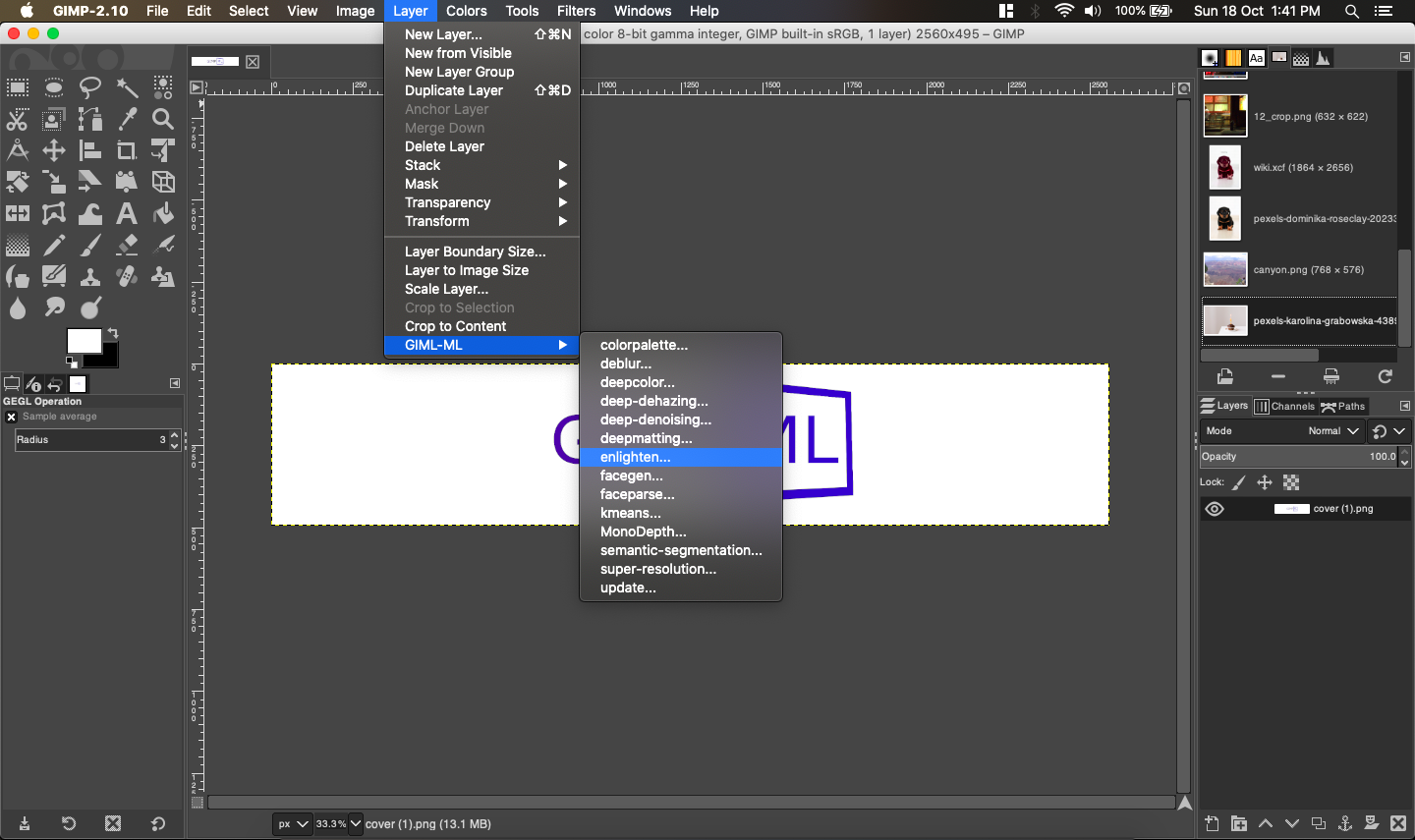}
	\caption{GIMP-ML Plugins Menu}
	\label{fig:screenshot}
	\end{figure}
	
	The rest of this paper is organized as follows. Section \ref{sec:dependencies} presents they key dependencies for GIMP-ML. This is followed by implementation details in Section \ref{sec:implementation}. Various applications of GIMP-ML have been illustrated in Section \ref{sec:applications}, which also includes links to demonstration videos on YouTube. Finally, conclusions and future work are presented in Section \ref{sec:conclusions}.
	
%
%
	
\section{Dependencies \label{sec:dependencies}} 

The Python package dependencies involved in the development of GIMP-ML are as follows:

\begin{enumerate}
    \item NumPy: The base N-dimensional array package, \texttt{numpy} \cite{walt2011numpy}, has been used for converting GIMP layer to a tensor for use in Pytorch.
    \item SciPy: The fundamental library for scientific computing, \texttt{scipy} \cite{jones2001scipy}, has been used for performing basic computing operations.
    \item OpenCV: The \texttt{opencv-python}\cite{mordvintsev2014opencv} package provides OpenCV libraries in Python. It has been used for edge detection.
    \item Pre-Trained Models: The \texttt{pretrainedmodels} includes a set of pre-trained models for PyTorch \cite{paszke2019pytorch}, of which the \texttt{InceptionResNetV2} has been used for the applications presented in this paper.
    \item Torch \& Torchvision: The \texttt{torch} \cite{paszke2019pytorch} and \texttt{torchvision} \cite{marcel2010torchvision} packages have been used to incorporate the deep learning framework through Pytorch.
\end{enumerate}

\section{Implementation Details \label{sec:implementation}} 
The GIMP-ML plugins have been developed in Python 2.7 which is supported in GIMP 2.10. A virtual environment has been separately created and added to the \texttt{gimp-python} path. This contains all the python packages used by the plugins. The plugins use CPU by default and switch to GPU for prediction when available. Additionally, there is an option to force the usage of CPU. The alpha channel in the input layer is dropped when not required as input to the deep learning network. The plugins take advantage of layers in GIMP for various workflows. As a consequence, image manipulation in the subsequent applications is also non-destructive in nature.

\section{Tools \label{sec:tools}} 
The tools available in GIMP-ML are summarized in Table \ref{tab:myfirstlongtable}. All deep learning based tools have a button to force the use of CPU.
\begin{longtable}{| p{.50\textwidth} | p{.50\textwidth} |} 
	
	\hline
	\centering
	\raisebox{-0.85\totalheight}{\includegraphics[width=0.3\textwidth]{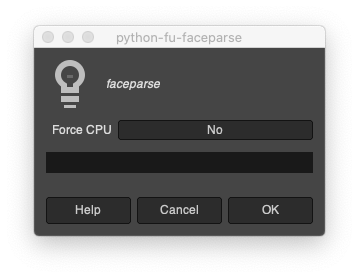}} & 
	\vfill
	\textbf{faceparse}: For segmenting portrait images, we used BiSeNet \cite{yu2018bisenet} trained on the CelebAMask-HQ dataset \footnote{\url{https://github.com/zllrunning/face-parsing.PyTorch}}. It can segment 19 classes such as skin, nose, eyes, eyebrows, ears, mouth, lip, hair, hat, eyeglass, earring, necklace, neck, and cloth. \\ 

	\hline
	\centering
	\raisebox{-0.85\totalheight}{\includegraphics[width=0.3\textwidth]{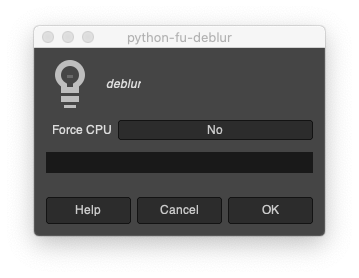}} & 
	\vfill
	\textbf{deblur}: We have ported the DeblurGAN-v2 \cite{kupyn2019deblurgan} model trained on the GoPro dataset \footnote{\url{https://seungjunnah.github.io/Datasets/gopro}} for deblurring images.\\ 

	\hline
	\centering
	\raisebox{-0.85\totalheight}{\includegraphics[width=0.3\textwidth]{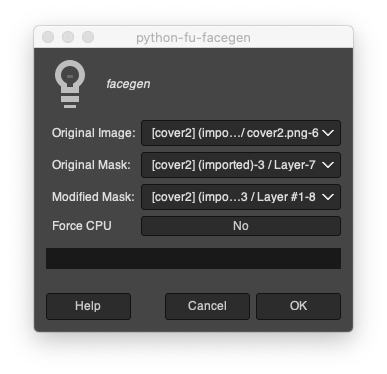}} & 
	\vfill
	\textbf{facegen}: With the \texttt{facegen} plugin, facial features in portrait photo can be segmented, modified and then newly generated. Trained on the CelebAMask-HQ dataset \cite{lee2019maskgan}, this model \footnote{\url{https://github.com/switchablenorms/CelebAMask-HQ}} relies on facial segmentation map generated in the previous faceparse sub-section. The mask can be duplicated into another layer and it can be manipulated using Color Picker Tool and Paintbrush Tool. The original input image , original mask and modified mask can then be fed into Mask-GAN to generate the desired image.\\ 

	\hline
	\centering
	\raisebox{-0.85\totalheight}{\includegraphics[width=0.3\textwidth]{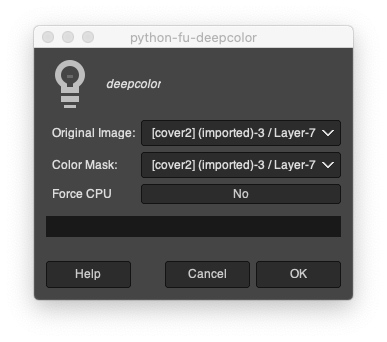}} & 
	\vfill
	\textbf{deepcolor}: User guided image colorization \footnote{\url{https://github.com/junyanz/interactive-deep-colorization}} as proposed by Zhang et. al. in \cite{zhang2017real} is available as the deepcolor tool. The model is trained on the ImageNet dataset \cite{deng2009imagenet}. The color mask layer should be transparent RGB layer (with alpha channel) that contains (local points) dots of size 6 pixels specifying which color should be present at which location. The tool can be used without this layer if the image and color mask layers are set to the same layer containing the image. \\

	\hline
	\centering
	\raisebox{-0.85\totalheight}{\includegraphics[width=0.3\textwidth]{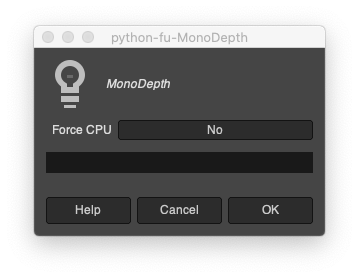}} & 
	\vfill
	\textbf{monodepth}: Disparity maps can be generated from images using deep learning methods and depth from stereo images. Recently, robust monocular depth estimation has been proposed in \cite{lasinger2019towards}. This model \footnote{\url{https://github.com/intel-isl/MiDaS}} has been ported for GIMP-ML. The model was trained using multi-objective optimization on several RGB-depth datasets.\\

	\hline
	\centering
	\raisebox{-0.85\totalheight}{\includegraphics[width=0.3\textwidth]{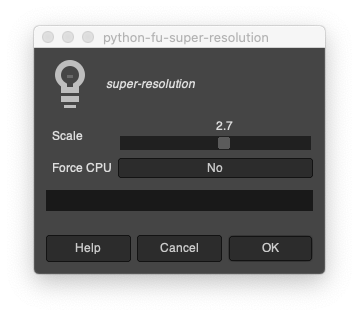}} & 
	\vfill
	\textbf{super-resolution}: The model in \cite{ledig2017photo} for image super resolution \footnote{\url{https://github.com/twtygqyy/pytorch-SRResNet}} was also implemented. Using this plugin the input image layer can be upscaled to upto 4x its original size.  \\

	\hline
	\centering
	\raisebox{-0.85\totalheight}{\includegraphics[width=0.3\textwidth]{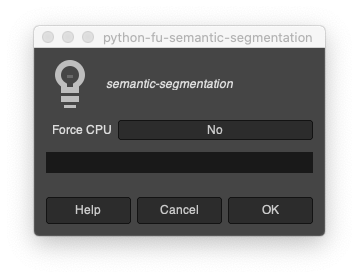}} & 
	\vfill
	\textbf{semantic-segmentation}: We used the Pytorch Deeplabv3 \cite{chen2017rethinking} model trained on the Pascal VOC dataset \cite{everingham2010pascal} for the semantic segmentation tool. It supports the 20 classes of Pascal VOC, namely, person, bird, cat, cow, dog, horse, sheep,  aeroplane, bicycle, boat, bus, car, motorbike, train, bottle, chair, dining table, potted plant, sofa, and tv/monitor. These objects can be directly segmented in images.  \\

	\hline
	\centering
	\raisebox{-0.85\totalheight}{\includegraphics[width=0.3\textwidth]{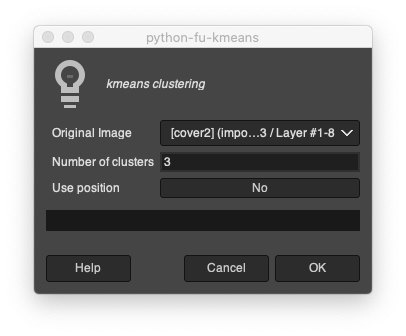}} & 
	\vfill
	\textbf{kmeans}: The scipy \cite{jones2001scipy} implementaion of kmeans was used. The tool requires the image layer, and number of clusters/colors in output. There is also an option to use (x,y) position coordinates as features for clustering.\\

	\hline
	\centering
	\raisebox{-0.85\totalheight}{\includegraphics[width=0.3\textwidth]{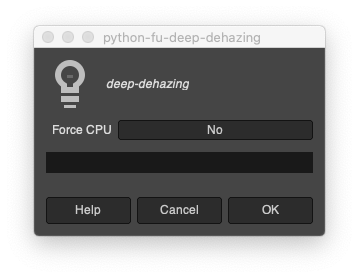}} & 
	\vfill
	\textbf{dehaze}: Image de-hazing based on deep learning \footnote{\url{https://github.com/MayankSingal/PyTorch-Image-Dehazing}} as proposed by Li et. al. in All-in-One Dehazing Network \cite{li2017aod} is available as the deep-dehaze tool . \\

	\hline
	\centering
	\raisebox{-0.85\totalheight}{\includegraphics[width=0.3\textwidth]{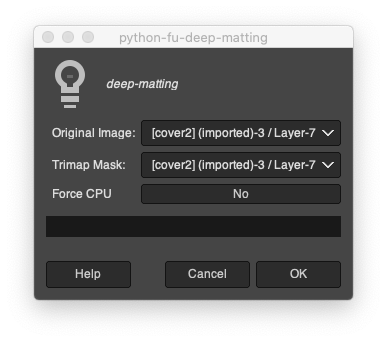}} & 
	\vfill
	\textbf{deepmatting}: Deep learning based image matting \footnote{\url{https://github.com/huochaitiantang/pytorch-deep-image-matting}} as proposed by Xu et. al. in \cite{xu2017deep} has also been ported into GIMP-ML. It requires two layers, namely the image layer and the trimap layer. The trimap Layer should contain segmentation mask of the object with RGB as [128,128,128] for boundaries, [255,255,255] for object the and [0,0,0] for background. \\

	\hline
	\centering
	\raisebox{-0.85\totalheight}{\includegraphics[width=0.3\textwidth]{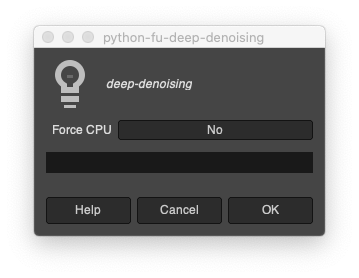}} & 
	\vfill
	\textbf{denoise}: Image de-noising \footnote{\url{https://github.com/yzhouas/PD-Denoising-pytorch}} as propsed by Zhou et.al. in \cite{zhou2019awgn} is available as the deep-denoise tool. \\

	\hline
	\centering
	\raisebox{-0.85\totalheight}{\includegraphics[width=0.3\textwidth]{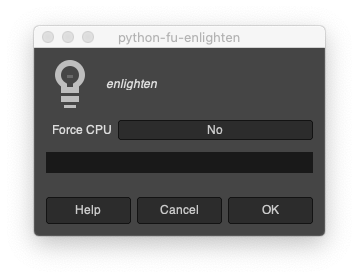}} & 
	\vfill
	\textbf{enlighten}: EnlightenGAN \footnote{\url{https://github.com/VITA-Group/EnlightenGAN}} proposed by Jiang et. al. in \cite{jiang2019enlightengan} is available as the enlighten tool.\\ \hline

	\caption{Tools available in GIMP-ML} 
	\label{tab:myfirstlongtable}
\end{longtable}



\section{Applications \label{sec:applications}} 
This section describes applications of GIMP-ML, which include background blurring, image re-coloring, face editing, generative portrait modification, monocular depth based relighting, motion blurring and generating super-resolution images. Demo videos of all the applications are available in the YouTube channel: \\ \url{https://youtube.com/user/kritiksoman}

\subsection{Background Blurring}
 The semantic-segmentation tool can be used to get object boundaries of the 20 predefined classes. We can then use it to selectively perform operations on regions of the image, such as blurring, hue/saturation change etc. A demonstration video for background blurring has been shown in \url{https://youtu.be/U1CieWi--gc}

\subsection{Image Re-coloring}
In order to re-colour an image, an RGB image can be converted to grayscale and then multiple colored version can be generated using different local hints layers. The resulting layers can then be selectively erased to retained colors as desired. An example has been shown in \url{https://youtu.be/4YpTa-gqEIw}.

\subsection{Face Editing}
 The segmentation map from faceparse tool can then be used to selectively manipulate various facial features. Hair color manipulation has been demonstrated in the video demo using this network. The demo for hair color manipulation using this plugin can be viewed at \url{https://youtu.be/thS8VqPvuhE}

\subsection{Generative Portrait Modification}
 Facegen tool can be used to modifiy facial features in a portrait image. A drawback of such a Mask-GAN is that it does not preserve unmodified facial features. This can, however, be taken care of by manually erasing unwanted facial feature changes from the generated layer thereby exposing the original image in the layer underneath. This is a valuable workflow since professional image editors spend a large amount of time in making portrait shots perfect and would retain original image facial features.
The demo has generative portrait modification has been shown in \url{https://youtu.be/kXYsWvOB4uk}

\subsection{Monocular Depth based Relighting}
Using the monodepth tool, the disparity map of images can be desaturated, inverted and colorized to created a layer representing light falling from the sky. In the demo video (\url{https://youtu.be/q9Ny5XqIUKk}), a day time image of a street has been converted to night time using this approach. As another example, a light painting has been created from a day time image and a tutorial has been shown in the youtube video \url{https://youtu.be/squyQYrllBg}.

\section{Conclusions and Future Work \label{sec:conclusions}}
This paper presented GIMP-ML, a set of Python plugins that enabled the use of deep learning models in GIMP via Pytorch for various applications. It has been shown that several manual and time-consuming image processing tasks can be simplified by the use of deep learning models, which makes it convenient for the users of image manipulation software to perform such tasks. GIMP 2.10 currently relies on Python 2.7 which been deprecated as on 1 January 2020. The next version of GIMP would use Python 3 and GIMP-ML codebase would be updated to support this. Further, deep learning models suffer from the data bias problem and only work well when the test image is from the same distribution as the data on which the model was trained. In future, the framework would be enhanced to handle such scenarios.

\bibliographystyle{unsrt}
\bibliography{references}
	
\end{document}